\documentclass[letterpaper, 10 pt, conference]{ieeeconf}  %
\IEEEoverridecommandlockouts                              
\usepackage[table]{xcolor}
\usepackage{graphicx} 
\usepackage{amsmath} 
\usepackage{amssymb}  
\usepackage{booktabs}
\usepackage{cite}
\usepackage{lipsum}
\usepackage{comment}
\usepackage{flexisym}
\usepackage{amsmath}
\usepackage{subcaption}
\usepackage{algpseudocode}
\usepackage{algorithm}
\usepackage{lipsum}
\usepackage{bm}
\usepackage{stfloats}
\usepackage{listings}
\usepackage{svg}
\usepackage{array}

\newcolumntype{L}{>{\centering\arraybackslash}m{3cm}}

\usepackage{multirow}
\usepackage{color, colortbl}

\usepackage{pifont}
\newcommand{\cmark}{\ding{51}}%
\newcommand{\xmark}{\ding{55}}%

\algnewcommand\algorithmicforeach{\textbf{for each}}
\algdef{S}[FOR]{ForEach}[1]{\algorithmicforeach\ #1\ \algorithmicdo}

\definecolor{Cyan}{rgb}{0.88,1,1}
\definecolor{Green}{rgb}{0.8,1,0.8}
\definecolor{Blue}{rgb}{0.80,0.89,1}

\newcommand{\fsnet}{\textit{FSNet}}

\newcommand{\etal}{\textit{et al.}}

\usepackage{subcaption}
\usepackage{cite}
\usepackage[normalem]{ulem}

\title{FSNet: A Failure Detection Framework for  Semantic Segmentation} 
\author{Quazi Marufur Rahman, Niko S{\"u}nderhauf, Peter Corke and Feras Dayoub %
\thanks{The authors are with Queensland University of Technology (QUT), Brisbane, Australia. We acknowledge the ongoing support from QUT's Centre for Robotics. Contact: {\tt\small quazimarufur.rahman@hdr.qut.edu.au}}%
}

\begin{document}

\maketitle

\begin{abstract}
Semantic segmentation is an important task that helps autonomous vehicles understand their surroundings and navigate safely. However, during deployment, even the most mature segmentation models are vulnerable to various external factors that can degrade the segmentation performance with potentially catastrophic consequences for the vehicle and its surroundings. To address this issue, we propose a failure detection framework to identify pixel-level misclassification. We do so by exploiting internal features of the segmentation model and training it simultaneously with a failure detection network. During deployment, the failure detector flags areas in the image where the segmentation model has failed to segment correctly. We evaluate the proposed approach against state-of-the-art methods and achieve $12.30\%$, $9.46\%$, and $9.65\%$ performance improvement in the AUPR-Error metric for Cityscapes, BDD100k, and Mapillary semantic segmentation datasets.

\end{abstract}

\section{Introduction}

Semantic segmentation using deep learning has become crucial for many safety-critical systems such as vision-based self-driving cars~\cite{treml2016speeding, Feng2021DeepMO} and robot-assisted surgery~\cite{allan20192017, Shvets2018AutomaticIS}. For instance, semantic segmentation is a significant component for any self-driving car for safety, reliability, and scene understanding \cite{Yang2018DenseASPPFS, hao2020brief}. Besides, it plays a substantial role in navigation ~\cite{maturana2018real, zhang2018road} and obstacle avoidance \cite{hua2019small, Arain2019ImprovingUO} by segmenting critical objects such as pedestrians and other vehicles in real-time from the visual sensory. Due to its importance, there is ongoing research \cite{sseg_lateef2019survey, sseg_Minaee2021ImageSU, GarciaGarcia2018ASO} to improve the overall performance of semantic segmentation to meet the safety-critical demand in robotic vision.

State-of-the-art research in semantic segmentation commonly assumes that the images encountered during training and later during deployment follow a similar distribution. However, this cannot be guaranteed for applications on autonomous vehicles that operate in the open, unconstrained world. The segmentation model will inevitably encounter situations (objects, object configurations, textures), environmental conditions (weather), or imaging conditions (motion blur, illumination, and exposure effects) that were never seen during training. As a result, a severe drop in segmentation performance could occur without prior warning, posing an extreme risk for the vehicle and its surroundings.

\begin{figure}
\begin{subfigure}{.49\columnwidth}
  \centering
  \includegraphics[width=\linewidth]{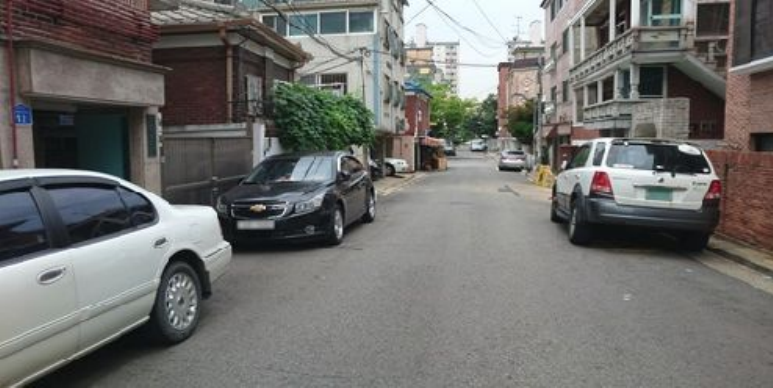}  
  \caption{}
  \label{fig:sub-first}
\end{subfigure}
\begin{subfigure}{.49\columnwidth}
  \centering
  \includegraphics[width=\linewidth]{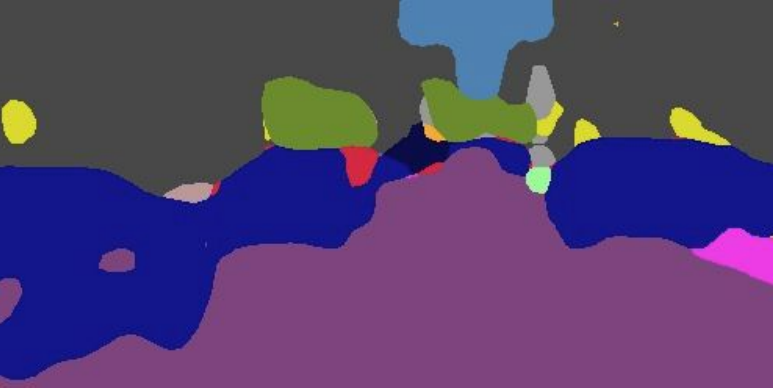}  
  \caption{}
  \label{fig:sub-second}
\end{subfigure}


\begin{subfigure}{.49\columnwidth}
  \centering
  \includegraphics[width=\linewidth]{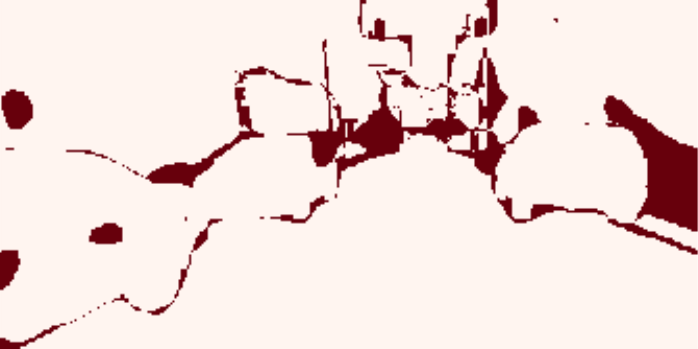}  
  \caption{}
  \label{fig:sub-third}
\end{subfigure}
\begin{subfigure}{.49\columnwidth}
  \centering
  \includegraphics[width=\linewidth]{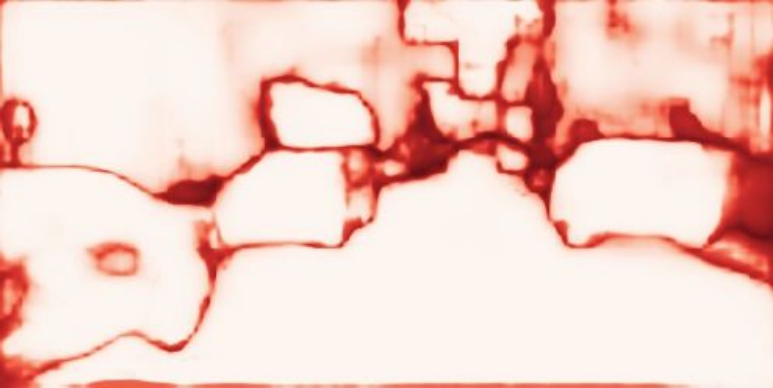}  
  \caption{}
  \label{fig:sub-fourth}
\end{subfigure}
\caption{A semantic segmentation network deployed on an autonomous vehicle may fail to predict the correct label in an input image for various reasons. Our proposed failure detection network is trained to identify the mismatch between prediction and ground truth. Here (a) and (b) respectively show input and output of the semantic segmentation network (c) A binary map showing the mismatch between ground truth and segmentation prediction. (d) Output from our failure detection network. The red colour highlights where the failure detection network identifies inaccurate prediction by the semantic segmentation network.}
\label{fig:hook}
\end{figure}

The ideal solution to achieve consistent semantic segmentation performance in all conditions is a highly effective, robust, and domain agnostic model trained using \emph{all} possible scenarios that it will encounter during the deployment phase. However, these requirements are infeasible for a practical scenario.  Another approach is to identify and remove inputs that can decrease segmentation accuracy. Out-of-distribution~\cite{ruff2021unifying} and open-set~\cite{geng2020recent} detection are examples of this approach. However, these approaches still can not scale to the complex semantic scenes and structure in which an autonomous vehicle operates~\cite{synthcp}. 

Similarly, uncertainty and confidence estimation can be used to detect incorrect semantic segmentation. However, recent works from \cite{synthcp} and \cite{9294308} show that these approaches alone are not effective enough to detect perception failure in semantic segmentation. Instead, \cite{zhang2014predicting, daftry2016introspective, synthcp} have argued in favor of using a specifically trained model to identify the incorrect perception of a target model without depending on approaches such as out-of-distribution, open-set, novelty detection, uncertainty, and confidence estimation.

In the context of semantic segmentation, several approaches such as failure prediction \cite{corbiere2019addressing, synthcp}, introspective perception \cite{zhang2014predicting, 9294308}, and quality prediction \cite{Devries2018LeveragingUE, robinson2018real, guruau2018learn}  train a separate model to identify the semantic segmentation failure. Most of these works are not applicable in autonomous vehicle scenarios, considering the complexity and significant variance of visual sensor inputs encountered by the semantic segmentation network during the deployment. Moreover, these works separately train the segmentation and the failure detection network from each other. Hence, the failure detection network can not access the potentially rich convolutional features of the segmentation network. Besides, a failure-dataset~\cite{synthcp, 9294308} is used to train the failure detection network. The failure-dataset is generated from a relatively smaller holdout set of the initial semantic segmentation training dataset. For example, \cite{9294308} uses $20\%$ of the segmentation dataset as failure-dataset to train the failure detection network. Therefore, the failure detector cannot exploit the available segmentation dataset and may lack generalizability.

This paper proposes a novel framework, \fsnet, consisting of semantic segmentation and a corresponding failure detection network. In contrast to existing works, both of these networks are connected and trained simultaneously. This architecture allows the failure detector to exploit the internal convolutional features of the semantic segmentation network, leading to better failure detection performance. Besides, \fsnet\ uses a joint learning technique to train both networks using the available semantic segmentation dataset. Hence, we can train the failure detection network using the entire segmentation dataset rather than relying on a failure-dataset. We evaluate \fsnet\ against the current SOTA methods using multiple datasets representing in and out-distribution scenarios. Although the segmentation and failure detection networks are connected and trained jointly, our framework does not impede the segmentation accuracy. Our experimental results show that the accuracy of our jointly trained segmentation network is similar to a separately trained network. At the same time, the jointly trained failure detection network outperforms all existing approaches. Figure~\ref{fig:hook} shows an example of semantic segmentation, the mismatch between the predicted segmentation and ground-truth, and how \fsnet\ can identify that incorrect segmentation. 

\section{Related Works}
\label{sec:literature review}

Failure detection or introspection is an essential requirement in robotics to ensure safety and reliability~\cite{rahman2021run}. Morris \cite{Morris-2007-9880} first proposed a \textit{robotic introspection} framework  to monitor operational state of robot for decision making purpose. Later ~\cite{triebel2016driven, grimmett2016introspective} extended this idea for semantic mapping and obstacle avoidance in robotics. Zhang~\etal~\cite{zhang2014predicting} introduced \textit{alert} -- a framework that predicts the failure of another model. A similar approach has been used by \cite{daftry2016introspective} for failure prediction in MAV, hardness predictor~\cite{wang2018towards} for image classifier, and performance monitoring~\cite{guruau2018learn} for robot perception system. In this work, we focus on detecting the failure of a semantic segmentation model in the autonomous vehicle context.

The study of failure detection or identifying the erroneous prediction of a model is closely related to uncertainty and confidence estimation. Hendrycks~\etal~\cite{hendrycks2016baseline} used Maximum Softmax Probability (MSP) derived from the softmax layer for detecting a failure in classification tasks. This work is considered as the standard baseline in related literature. However, MSP suffers drawbacks such as failure to distinguish between in and out-distribution samples and improper calibration. To reduce the risk of making incorrect classification, Geifman~\etal~\cite{geifman2017selective} introduced \textit{selective classifier}. This approach controls and guarantees the risk level of a classifier by using thresholds on pre-defined confidence functions, e.g., MSP. Heinrich~\etal~\cite{Jiang2018ToTO} introduced \textit{trust score} that compares the prediction between a classifier and a modified nearest neighbor classifier to measure classifier reliability. Most recently, Corbiere~\etal~\cite{corbiere2019addressing} has presented \textit{true class probability} to improve the unreliable ranking of confidence score. Besides, MC Dropout based techniques have become popular for failure detection in classification. However, Xia~\etal~\cite{synthcp} has argued that these approaches are not applicable in semantic segmentation because of the lack of information on semantic structure and contexts.

Failure detection in the context of semantic image segmentation is being studied extensively in recent years. Kohlberger~\etal~\cite{Kohlberger2012EvaluatingSE} used a novel space of segmentation features to predict overlap error and the Dice coefficient of an organ segmentation model. Later, Valindria~\etal~\cite{Valindria2017ReverseCA} have introduced \textit{reverse classification accuracy} to predict segmentation quality of medical image segmentation. Huang~\etal~\cite{Huang2016QualityNetSQ} showed that segmentation quality could be predicted using~\textit{QualityNet}. ~\cite{Jungo2018UncertaintydrivenSC, Devries2018LeveragingUE} showed the application of Bayesian CNN for predicting semantic segmentation failure. ~\cite{chabrier2006unsupervised, gao2017novel} used unsupervised learning to quantify the quality of semantic segmentation tasks. However, because providing image-level segmentation quality rather than pixel-level failure detection, these works do not apply to identifying the areas where semantic segmentation is incorrect.

Xia~\etal~\cite{synthcp} have introduced \textit{SynthCP} to predict pixel-level failure in semantic segmentation. They also demonstrated the usage of~\cite{hendrycks2016baseline, corbiere2019addressing, gal2016dropout, 9294308} for a similar task. Here,~\cite{synthcp} and~\cite{9294308} do not access the internal convolutional features of the segmentation network during the training or inference phase. Besides, these works explicitly use a failure-dataset to train the failure detector and consequently can not exploit the entire segmentation dataset. \cite{hendrycks2016baseline, gal2016dropout, corbiere2019addressing} use indirect approaches of using segmentation confidence and entropy for failure detection, which is suboptimal \cite{synthcp, 9294308}.

Our proposed framework \fsnet\ focuses on addressing the shortcomings of the current research. Firstly, this framework enables the failure detector to access the convolutional features of the segmentation network. Secondly, \fsnet\ jointly trains its networks using the entire segmentation dataset without requiring the explicit failure-dataset. Both of these techniques contribute to the better performance of \fsnet\ than all other existing approaches.

\section{Approach Overview}
\label{sec:approach overview}
This section introduces our failure detection framework \fsnet\ for semantic segmentation. \fsnet\ consists of two connected components -- one semantic segmentation network and one failure detection network. We will describe both of these networks and how they work jointly to detect the failure of semantic segmentation.

\subsection{Module Architecture}


\begin{table}[b]
\centering
\caption{List of layers to preprocess the input for $F_{E1}$ and $F_{E2}$ -- the two encoders of the failure detection network $F$. Each layer generates one channel output from segmentation logits $l$ without changing input width and height. }
\label{tbl:layer infos}
\renewcommand{\arraystretch}{1.5}

\begin{tabular}{l|l|l}
\hline
Layer   & Operation              & Description                                                                                                                           \\ \hline
$\mathcal{W}_{1}(l)$ & \textit{Conv1x1(l)} & \multicolumn{1}{m{4cm}}{Applies \textit{2D} convolution on $l$ with kernel size $1$.}\\
$\mathcal{W}_{2}(l)$ & \textit{Max(Softmax(l))} & \multicolumn{1}{m{4cm}}{Extracts maximum softmax value across each channel of $l$.}\\
$\mathcal{W}_{3}(l)$ & \textit{Sigmoid(Entropy(l))} & \multicolumn{1}{m{4cm}}{ Uses sigmoid normalization after getting channelwise entropy of $l$.}\\
$\mathcal{W}_{4}(l)$ & \textit{ArgMax(l)} & \multicolumn{1}{m{4cm}}{Returns indices of the maximum value across each channel of $l$. Same as segmentation label.}\\ 
\hline
\end{tabular}%

\end{table}

\fsnet\ uses a joint architecture to connect the semantic segmentation and corresponding failure detection network and trains these networks end-to-end using the semantic segmentation dataset. \fsnet\ also allows the failure detection network to access internal features of the segmentation network during training and inference.

Let $S$ be a basic semantic segmentation network combining a convolutional encoder $S_{E}$ and decoder $S_{D}$. $S$  classifies each pixel of a given image $x$ of shape $w \times h \times 3$ into a particular label from a set $\mathcal{C} = \{1,2, \ldots, C\}$. $S_{D}$ uses the convolutional features $e$ from the last layer of $S_{E}$ to generate logits $l$ of size $w \times h \times C$. Here $e=S_{E}(x)$ and $l = S_{D}(e)$. Based on architectural choice, $S_{D}$ may exploit features from different layers of $S_{E}$. Later, a softmax function is applied on $l$ to generate the predicted label map $\hat{y} = S(x) \in \mathcal{C} ^ {w \times h}$.

\begin{figure}[]
 \centering
 \includegraphics[width=0.99\columnwidth]
 {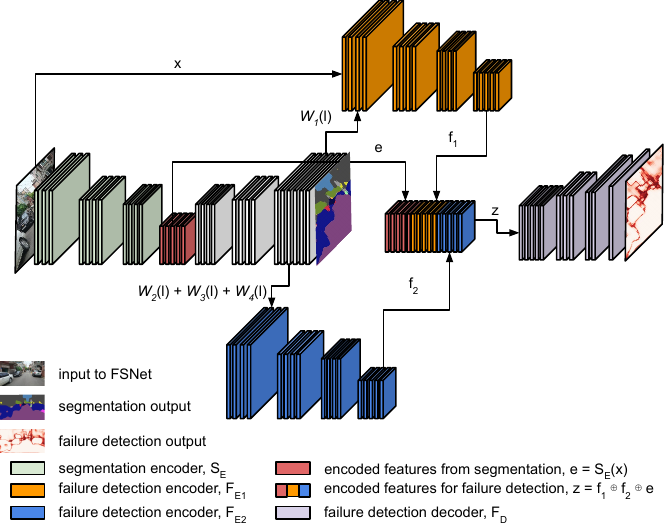}
 \caption{An outline of \fsnet\ framework, consisting of the semantic segmentation and failure detection network. It shows the inter-connection between the segmentation and failure detection network.}
\label{fig:ral_architecture}
\end{figure}

We are proposing a failure detection network $F$ to predict $\bar{y}$, a $2D$ failure map of size $w \times h$ indicating the pixels where $\hat{y}$ is incorrect. $F$ consists of two encoders $F_{E1}$ and $F_{E2}$, and a decoder $F_{D}$. These encoders take $l$ as input after preprocessed by four different layers listed in Table~\ref{tbl:layer infos}. First, $F_{E1}$ uses $x \oplus \mathcal{W}_{1}(l)$ as its input and produces the encoded feature $f_{1}$. Meanwhile, $F_{E2}$ generates encoded feature $f_{2}$ from $\mathcal{W}_{2}(l)  \oplus \mathcal{W}_{3}(l) \oplus \mathcal{W}_{4}(l)$. Here, $\oplus$ represents channel-wise concatenation operation. Later, using Equation~\ref{eqn:fx3}, $f_{1}$, $f_{2}$ and $e$ are concatenated to generate feature $z$ for the failure detection decoder $F_{D}$.
\begin{equation}
\label{eqn:fx3}
    z = f_{1} \oplus f_{2} \oplus e.
\end{equation}
$F_{D}$ takes $z$ as input and upsamples it to generate the failure map $\tilde{y}$ of size $w \times h$. $\tilde{y}_{ij}$ represents the confidence of $F_{D}$ that $S$ has misclassified $x_{ij}$. Figure~\ref{fig:ral_architecture} shows an overview of \fsnet\ and the inter-connection among its different components.

\subsection{Training Procedure}

We use a single dataset and two different loss functions -- segmentation loss and failure detection loss -- to jointly train the semantic segmentation network $S$ and the failure detection network $F$ of \fsnet. Let, for each input $x$, $S$ predict the label as $\hat{y}$. The cross-entropy loss function in Equation~\ref{eqn:cross-entropy} is used to calculate the segmentation loss $L_{1}$ from ground-truth $y$ and label prediction $\hat{y}$. \begin{equation}
    \label{eqn:cross-entropy}
    L_{1} = -\sum_{i}^{C} y_{i} \log (\hat{y}_{i})
\end{equation}

The task of $F$ is to predict the failure of $S$. Therefore, a ground truth $\bar{y}$ is required to represent that failure, so that we can train $F$. Here, $\bar{y}$ is a $2D$ binary label of size $w \times h$. Each pixel of $\bar{y}$ is either \emph{one} or \emph{zero}, indicating the difference or similarity between $y$ and $\hat{y}$. We apply Equation~\ref{eqn:failure_map} to generate this failure detection ground-truth $\bar{y}$. During training, $F$ is optimized to predict the $\bar{y}$. Assuming $\tilde{y}$ as the output of the failure detection network, we use the balanced binary cross-entropy loss of Equation~\ref{eqn:binary-cross-entropy} to calculate the failure detection loss $L_{2}$ where $\beta$ is the weight balancing factor.

\begin{equation}
\label{eqn:failure_map}
    \bar{y}_{ij} = \begin{cases}
      1 & y_{ij} \neq \hat{y}_{ij}\\
      0 & \text{otherwise}
    \end{cases}
\end{equation}

\begin{equation}
\label{eqn:binary-cross-entropy}
    L_{2} = -(\beta \cdot \bar{y}\log(\tilde{y}))+(1-\bar{y})\log(1-\tilde{y})
\end{equation}

There are two steps in the training procedure. At first, we backpropagate only the loss $L_{1}$ into \fsnet\ until the segmentation network $S$ is converged. This step only trains $S$ to perform semantic segmentation. After the convergence of $S$, $L_{1} + L_{2}$ is used as the new loss. As $S$ and $F$ are connected, this step jointly optimizes semantic segmentation and failure detection networks. Our experiment shows that without converging $S$ first, we can not jointly optimize both networks of \fsnet.

\section{Experimental Setup}
\label{sec:experimental setup}
This section describes the experimental setup used for \fsnet\ evaluation. First, we will discuss the in- and out-distribution dataset settings. Later, existing approaches, evaluation metrics, and implementation will be detailed.

\textbf{In- and Out-Distribution Dataset.} We used a training dataset of 2974 images from the Cityscapes \cite{cordts2016cityscapes} dataset to train \fsnet\ and other approaches. There are two settings in the experimental setup. The first one is the in-distribution, where the training and testing data come from the same distribution. We used a testing dataset of 500 images from the Cityscapes for evaluation. In the out-distribution setting, the testing data comes from a different dataset. We used 1000 images from the BDD100k \cite{yu2018BDD100k} and randomly selected 1000 images from the Mapillary \cite{neuhold2017mapillary} semantic segmentation dataset. All the segmentation networks are trained to segment the 19 classes available in the Cityscapes dataset.

\textbf{Methods to Compare} We compare \fsnet\ failure detection network to multiple methods -- MSP~\cite{hendrycks2016baseline}, MCDropout~\cite{gal2016dropout}, TCP~\cite{corbiere2019addressing}, Direct-prediction~\cite{9294308} and SynthCP~\cite{synthcp}. MSP and MCDropout are standard baselines for pixel-level failure detection. Direct-prediction, TCP, and SynthCP use a separate failure-dataset to train the failure detection network. Using this new dataset  Direct-prediction uses a separate network to train their failure detector. TCP trains a network to predict the true class probability as a failure indicator. Most recently, SynthCP used a conditional GAN and a comparison module to train a model that identifies the failure of semantic segmentation. SynthCP is the SOTA approach among these works. 

In contrast to existing approaches, \fsnet\ jointly trains the semantic segmentation and the corresponding failure detection network. Although jointly trained, \fsnet\ segmentation network should perform similarly to the individually trained segmentation network. To ensure this, we will compare \fsnet\ segmentation network accuracy with the individually trained SynthCP segmentation network.

\begin{table}[]
\centering
\caption{Semantic segmentation accuracy in mIOU and Cls-Acc metrics. It shows the performance of individually trained SynthCP and jointly training \fsnet\ segmentation model performance in three different datasets. Cityscapes dataset is used to train both approaches.}
\label{tbl:seg_result_table}
\begin{tabular}{@{}cc|cc|cc@{}}
\hline
\multirow{2}{*}{Dataset} & \multirow{2}{*}{Methods} & \multicolumn{2}{c}{FCN8} & \multicolumn{2}{c}{DeepLabV2} \\ \cline{3-6} 
                            &          & mIOU$\uparrow$   & Cls-Acc$\uparrow$ & mIOU$\uparrow$   & Cls-Acc$\uparrow$ \\ \hline
Cityscapes & SynthCP  & 29.84 & 35.19  & 32.98 & 37.76  \\
                        {\scriptsize in-distribution}    & \fsnet\ & \textbf{29.92} & \textbf{39.23}  & \textbf{33.22} & \textbf{41.70}  \\ \hline
BD100k     & SynthCP  & 19.42 & 24.52  & \textbf{23.97} & 28.46  \\
      {\scriptsize out-distribution}                      & \fsnet\ & \textbf{19.55} & \textbf{25.95}  & 23.54 & \textbf{30.25}  \\ \hline
      
Mapillary  & SynthCP  & 18.90 & 25.28  & \textbf{23.40} & 29.05  \\
         {\scriptsize out-distribution}                   & \fsnet\ & \textbf{19.32} & \textbf{26.97}  & 23.01 & \textbf{30.76} \\
                            \hline
\end{tabular}
\end{table}

\textbf{Evaluation Metrics.} Following \cite{corbiere2019addressing, synthcp}, we used AP-Err, AP-Suc, AUC, and FPR95 as evaluation metrics. AP-Err considers incorrect prediction as positive class and computes the Area Under the Precision-Recall (AUPR) curve. AP-Suc computes AUPR too but considers correct prediction as to the positive class. AUC calculates the area under the Receiver Operating Characteristics, and FPR95 computes False-Positive Rate at 95\% True-Positive Rate. \fsnet\ segmentation network is compared with the SynthCP segmentation network using mean Intersection over Union (mIOU) and per-class accuracy (Cls-Acc). mIOU first calculates the IOU for each class and then calculates the average over classes. Cls-Acc measures the percentage of correctly labeled pixels for each semantic class and then averages over the classes.

\begin{table*}[t]
\caption{Failure detection experiments on the Cityscapes, BDD100k, and Mapillary dataset using AP-Err, AP-Suc, AUC, and FPR95 metrics. All approaches are trained using the Cityscapes training dataset to detect the failure of FCN8 and DeepLabV2 semantic segmentation networks. Test dataset Cityscapes refers to in-distribution and BDD100k, Mapillary refers to out-distribution settings.}
\centering
\resizebox{\textwidth}{!}{%
\begin{tabular}{ll|cccc|cccc}
\hline
\multirow{2}{*}{Test Dataset} & \multirow{2}{*}{Methods} & \multicolumn{4}{c}{FCN8} & \multicolumn{4}{c}{DeepLabV2} \\ \cline{3-10} 
               & & AP-Err$\uparrow$ & AP-Suc$\uparrow$ & AUC$\uparrow$   & FPR95$\downarrow$ & AP-Err$\uparrow$ & AP-Suc$\uparrow$ & AUC$\uparrow$   & FPR95$\downarrow$ \\ \hline
\multirow{6}{*}{\begin{tabular}[x]{@{}c@{}}Cityscapes\\ \scriptsize in-distribution\end{tabular}}  & SynthCP\cite{synthcp}  & 55.53          & \textbf{99.18} & 92.92          & 22.47          & 49.99          & \textbf{99.34} & 92.98          & 21.69          \\
 & MCDropout\cite{gal2016dropout}        & 49.23  & 99.02  & 91.47 & 25.16 & 47.85  & 99.23  & 92.19 & 24.68 \\
 & MSP\cite{hendrycks2016baseline}               & 50.31  & 99.01  & 91,54 & 25.34 & 48.46  & 99.24  & 92.26 & 24.41 \\
 & Direct\cite{9294308}           & 52.16  & 99.14  & 92.55 & 22.34 & 48.76  & 99.34  & 92.94 & 21.56 \\
 & TCP\cite{corbiere2019addressing}               & 48.54  & 98.82  & 90.29 & 32.20 & 45.57  & 98.84  & 89.14 & 36.98 \\
 & \fsnet\ (Ours) & \textbf{67.83} & 98.98          & \textbf{94.35} & \textbf{21.52} & \textbf{57.84} & 99.13          & \textbf{93.97} & \textbf{21.39} \\ \hline

 \multirow{6}{*}{\begin{tabular}[x]{@{}c@{}}BDD100k\\\scriptsize out-distribution \end{tabular}}  & SynthCP~\cite{synthcp}               & 62.83  & 84.44 & 76.58 & 64.21 & 53.60  & 88.46 & 75.71 & 61.55 \\
& MCDropout~\cite{gal2016dropout}             & 59.77  & 83.78          & 74.85 & 64.68 & 38.26  & 84.18          & 66.27 & 70.10 \\
& MSP~\cite{hendrycks2016baseline}                   & 60.40  & 83.65          & 75.04 & 65.14 & 54.74  & 88.72 & 76.32 & 63.06 \\
& Direct~\cite{9294308}                & 61.77  & 84.91 & 76.85 & 62.51 & 53.18  & 88.28          & 75.53 & 61.81 \\
& TCP~\cite{corbiere2019addressing}                   & 57.99  & 77.54          & 70.77 & 75.66 & 47.9   & 83.02          & 67.47 & 75.10 \\
& \fsnet\ (Ours)                               & \textbf{72.29} & \textbf{86.63} & \textbf{83.17} & \textbf{51.81} & \textbf{71.81} & \textbf{91.76} & \textbf{86.94} & \textbf{43.56} \\ \hline

 \multirow{6}{*}{\begin{tabular}[x]{@{}c@{}}Mapillary\\\scriptsize out-distribution\end{tabular}}  & SynthCP~\cite{synthcp}           & 58.64  & 92.97 & 82.22 & 46.91 & 49.51  & 93.72  & 79.57 & 47.05 \\
& MCDropout~\cite{gal2016dropout}         & 57.34  & 93.30          & 82.11 & 46.81 & 47.39  & 93.49  & 78.48 & 52.67 \\
& MSP~\cite{hendrycks2016baseline}               & 58.29  & 92.97          & 82.53 & 45.35 & 48.03  & 93.65  & 79.32 & 46.63 \\
& Direct~\cite{9294308}            & 55.50  & 90.90          & 79.30 & 53.44 & 43.79  & 92.57  & 75.03 & 63.33 \\
& TCP~\cite{corbiere2019addressing}               & 56.21  & 93.23          & 81.77 & 47.19 & 32.97  & 92.88  & 73.61 & 55.43 \\
& \fsnet\ (Ours) & \textbf{68.29} & \textbf{93.76} & \textbf{87.75} & \textbf{40.19} & \textbf{63.19} & \textbf{95.68} & \textbf{88.95} & \textbf{37.36} \\ \hline

\end{tabular}%
\label{tbl:fail_result_table}
}
\end{table*}

\textbf{Implementation.} Following~\cite{synthcp}, we used FCN8 \cite{fcn_long2015fully} and DeepLabV2 \cite{chen2017deeplab} as the semantic segmentation networks  in our framework. FCN8 and DeeplabV2 are based on VGG16 and ResNet101 backbone networks, respectively, and pretrained on the MS-COCO semantic segmentation dataset. Both encoders in the \fsnet\ failure detection network use the ResNet18 network pretrained on the ImageNet dataset. Our training process consists of two steps. At first, the \fsnet\ segmentation network is trained only using the segmentation loss for $20k$ iterations for convergence. For this step, we follow the hyper-parameters and image augmentations used by SynthCP. Then, for the next $10k$ iterations, \fsnet\ is trained using segmentation and failure detection loss. The failure detection network uses \textit{adam} optimizer with learning rate $0.0002$.

\section{Evaluation}
\label{sec: evaluation and results}

This section evaluates the semantic segmentation and failure detection accuracy of \fsnet\ with the existing approaches. It also shows comparative performance for in-distribution and out-distribution settings.

\subsection{Semantic Segmentation Evaluation} 
Table~\ref{tbl:seg_result_table} shows the semantic segmentation accuracy between SynthCP and \fsnet. In the in-distribution setting, the segmentation accuracy of \fsnet\ based on FCN8 and DeepLabV2 improves by $4.0\%$ in the Cls-Acc metric. \fsnet\ also shows better performance in the mIOU metric. Both SynthCP and \fsnet\ demonstrated lower accuracy than the in-distribution settings as BDD100k, and Mapillary datasets were unknown to the segmentation network. However, \fsnet\ segmentation accuracy is better than SynthCP for out-distribution setting too. This result shows that we can train segmentation and failure detection networks jointly without degrading the segmentation accuracy.

\begin{figure*}[t]
\centering
\begin{subfigure}{0.33\textwidth}
\centering
    \includegraphics[width=.99\textwidth]{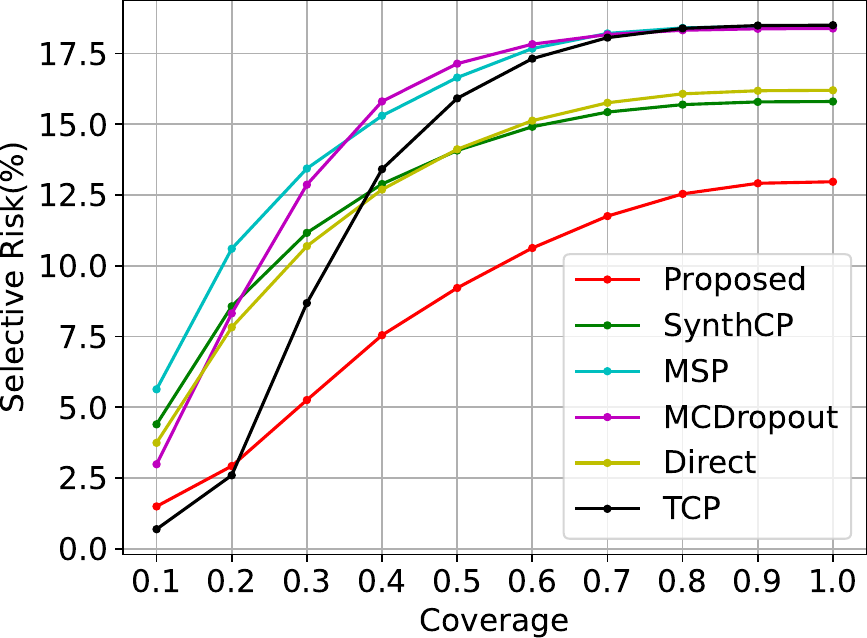}
    \caption{}
    \label{fig:risk_coverage_model_fcn8_in_cityscapes_out_cityscapes}
\end{subfigure}%
\begin{subfigure}{0.33\textwidth}
\centering
    \includegraphics[width=.99\textwidth]{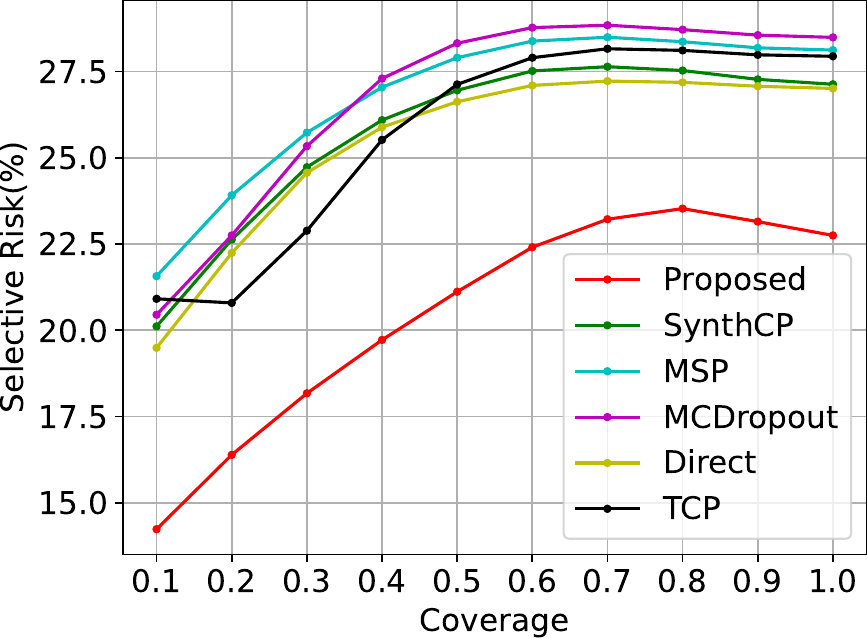}
    \caption{}
    \label{fig:risk_coverage_model_fcn8_in_cityscapes_out_bdd}
\end{subfigure}%
\begin{subfigure}{0.33\textwidth}
\centering
    \includegraphics[width=.99\textwidth]{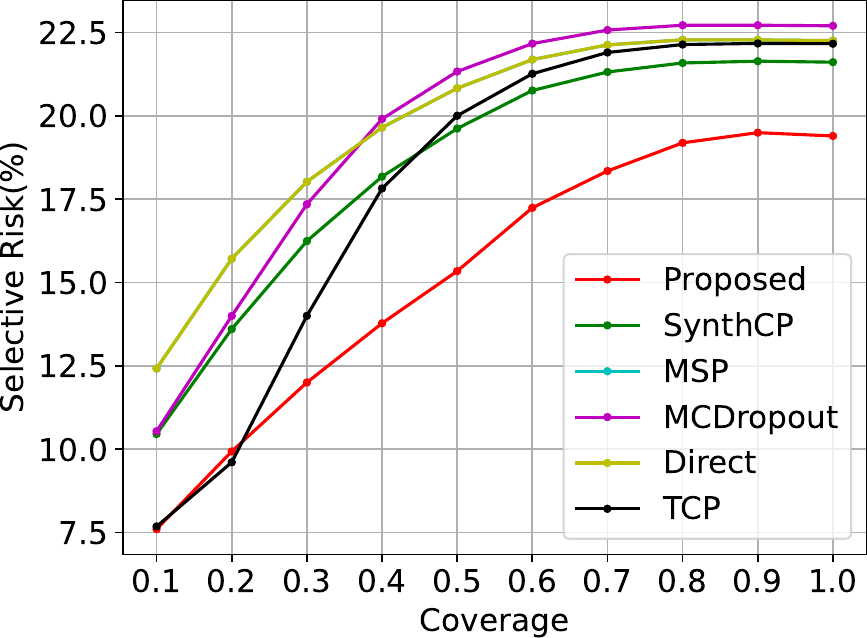}
    \caption{}
    \label{fig:risk_coverage_model_fcn8_in_cityscapes_out_mapillary}
\end{subfigure}

\begin{subfigure}{0.33\textwidth}
\centering
    \includegraphics[width=.99\textwidth]{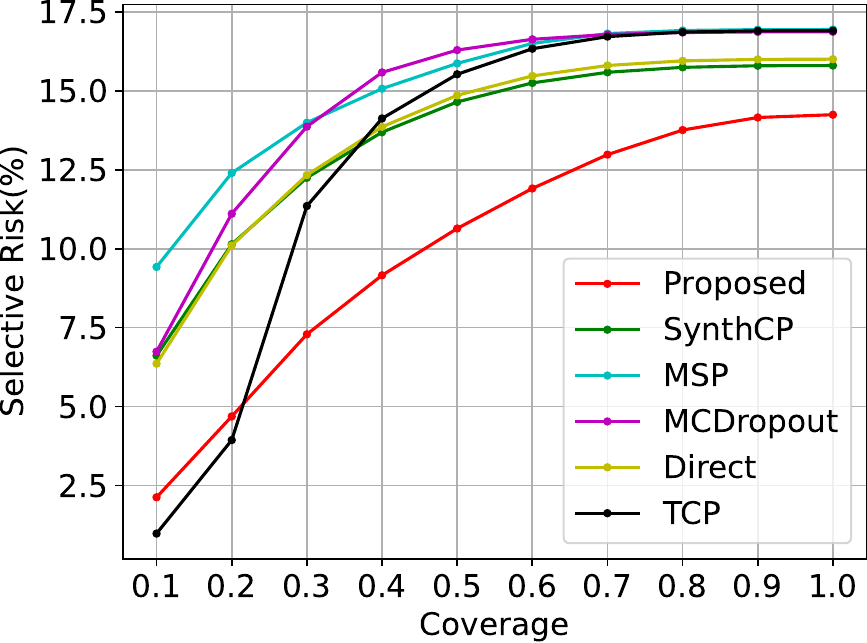}
    \caption{}
    \label{fig:risk_coverage_model_deeplab_in_cityscapes_out_cityscapes}
\end{subfigure}%
\begin{subfigure}{0.33\textwidth}
\centering
    \includegraphics[width=.99\textwidth]{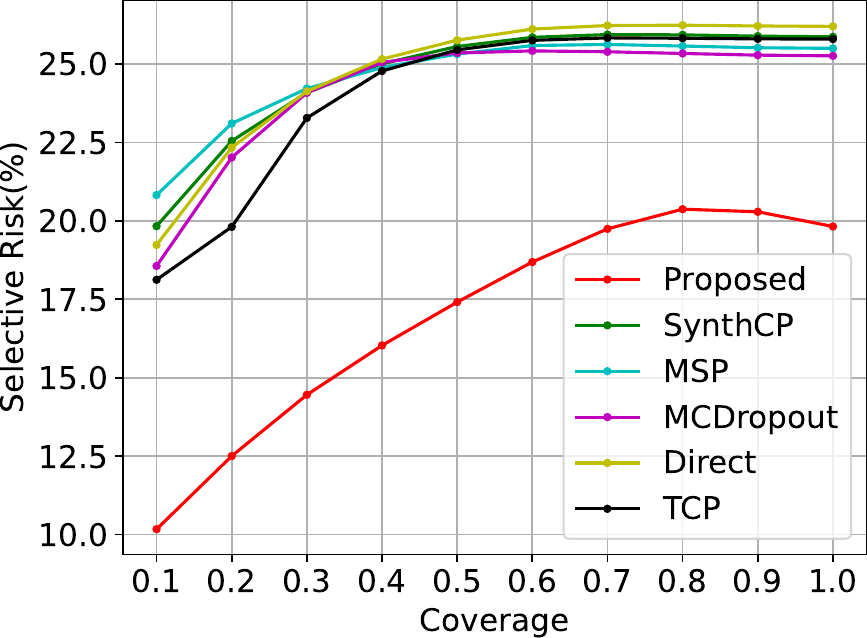}
    \caption{}
    \label{fig:risk_coverage_model_deeplab_in_cityscapes_out_bdd}
\end{subfigure}%
\begin{subfigure}{0.33\textwidth}
\centering
    \includegraphics[width=.99\textwidth]{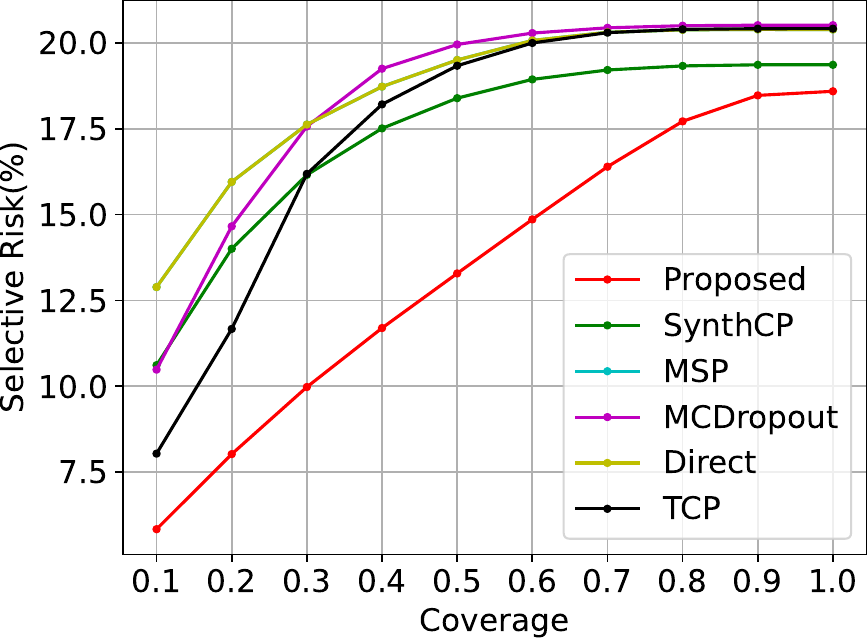}
    \caption{}
    \label{fig:risk_coverage_model_deeplab_in_cityscapes_out_mapillary}
\end{subfigure}
\caption{The Risk-Coverage curve for semantic segmentation failure detection. (a), (b) and (c) illustrate the Risk-Coverage curve for FCN8 failure detection in Cityscapes, BDD100k, and Mapillary datasets. (d), (e) and (f) demonstrate the Risk-Coverage curve for DeepLabV2 in the same dataset settings. Here (a) and (d) represent the in-distribution and (b), (c), (e), and (f) represent the out-distribution setting. In all cases, \fsnet\ has a lower risk of failure for all coverage levels.}
\label{fig:risk_coverage_fcn8}
\end{figure*}

\subsection{Failure Detection Evaluation}

Table~\ref{tbl:fail_result_table} shows the failure detection accuracy of \fsnet\ and all existing approaches using AP-Err, AP-Suc, AUC, and FPR95 metrics. These metrics are averaged over 19 classes of the Cityscapes dataset. For the in-distribution setting, \fsnet\ failure detection network achieves $67.83$ in AP-Err for identifying failure of FCN8, which is $12.30\%$ higher than the SOTA -- SynthCP. It also outperforms SynthCP in  AUC and FPR95. However, \fsnet\ is $0.2\%$ inferior to SynthCP in the AP-Suc metric. In contrast to SynthCP, \fsnet\ uses a balanced binary cross-entropy loss function to train the failure detection network. It has improved \fsnet\ AP-Err by a large margin for a negligible performance reduction in the AP-Suc. In the same settings, \fsnet\ for DeepLabV2 demonstrates a similar trend by outperforming SynthCP in the AP-Err metric by $7.85\%$.

Table~\ref{tbl:fail_result_table} also shows  \fsnet\ failure detection accuracy for FCN8 and DeepLabV2 in out-distribution setting. Here, we trained \fsnet\ using the Cityscapes dataset and evaluated using the BDD100k and Mapillary datasets. These experiments illustrate the \fsnet\ generalization capability. In four metrics and two datasets, \fsnet\ outperforms all the existing methods. 

We see higher AP-Err in out-distribution than in the in-distribution setting. The reason is the lower semantic segmentation accuracy for out-distribution (see Table~\ref{tbl:seg_result_table}). As the segmentation networks make more label prediction errors in the out-distribution setting, the failure detection network can identify these errors and hence shows better performance in out-distribution than in-distribution.

\begin{figure*}[t]
    \centering
    \includegraphics[width=0.98\textwidth]{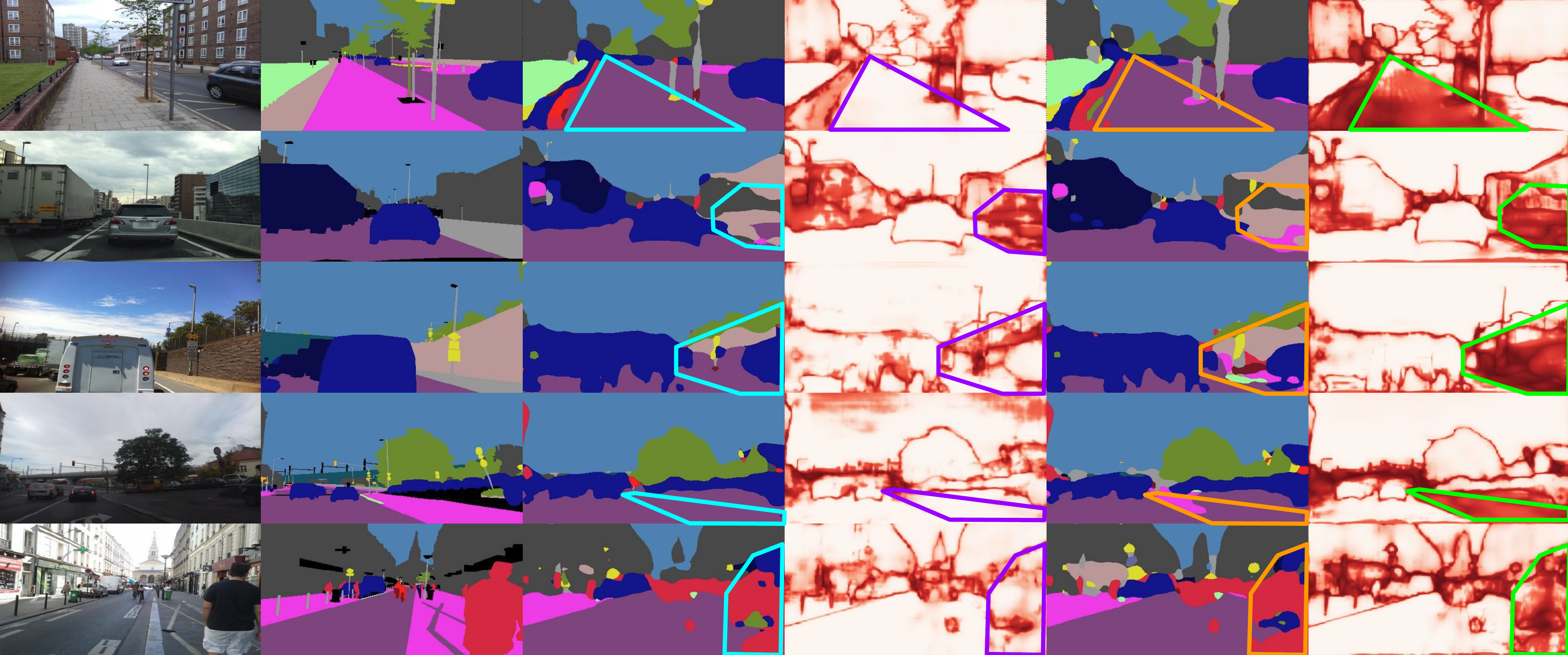}
    \caption{Semantic segmentation and failure detection example of SynthCP and \fsnet. $1^{st}$  column shows the input images to both approaches. $2^{nd}$ column is the ground-truth segmentation label. $3^{rd}$ column shows the segmentation by SynthCP, and Cyan boxes show the incorrectly segmented areas. $4^{th}$ column is the segmentation failure detected by SynthCP, and here  Purple boxes highlight the incorrectly detected segmentation failure by SynthCP. $5^{th}$ column shows the segmentation by \fsnet. Orange boxes demonstrate the area where \fsnet\ segmentation is incorrect. $6^{th}$ column is the \fsnet\ failure map. Here the green boxes highlight the area where \fsnet\ detects failure better than the SynthCP.}
    \label{fig:examples}
\end{figure*}
\subsection{Risk-Coverage Evaluation} 
We use the Risk-Coverage~\cite{geifman2017selective} curve to evaluate the impact after detecting semantic segmentation failure. Here, Coverage is the percentage of predicted pixel labels that are not flagged as a failure by \fsnet, and Risk is the percentage of misclassification error in those predictions. Based on this metric, \fsnet\ can reject the segmentation network's prediction to achieve the desired risk level.

Figure~\ref{fig:risk_coverage_model_fcn8_in_cityscapes_out_cityscapes} shows Risk-Coverage curves for all approaches while detecting the failure of FCN8 on Cityscapes dataset for in-distribution setting. We plot these curves using ten different Coverage levels. Here \fsnet\ demonstrates lower risk than all existing approaches. As an example, for $60\%$ coverage, $10.5\%$ risk means \fsnet\ has rejected $40\%$ segmentation prediction assuming that the prediction is incorrect, keeping the coverage $60\%$. In this coverage, $10.5\%$ pixels have been incorrectly classified. All other methods show risk levels from $15\%$ to $17.5\%$. For DeepLabV2 and in-distribution setting, \fsnet\  risk of failure is $12\%$ for $60\%$ coverage while other existing approach's risk varies from $15\%$ to $16.5\%$.

Figure~\ref{fig:risk_coverage_model_fcn8_in_cityscapes_out_bdd} and Figure~\ref{fig:risk_coverage_model_fcn8_in_cityscapes_out_mapillary} show risk-coverage curves for FCN8 in out-distributions setting. In both cases, \fsnet\ show lower risk level than all exisiting methods for all coverage levels. Figure~\ref{fig:risk_coverage_model_deeplab_in_cityscapes_out_bdd} and Figure~\ref{fig:risk_coverage_model_deeplab_in_cityscapes_out_mapillary} show risk-coverage curve for DeepLabV2 in out-distribution settings with the similar trend where \fsnet\ outperforms all existing approaches.

Figure~\ref{fig:examples} shows qualitative results and the comparison between \fsnet\ and SynthCP for detecting the failure of semantic segmentation.

\subsection{Discussion}

\begin{table*}[t]
\caption{We have trained a multi-branch failure detection network using the entire segmentation dataset and removed one feature at a time to study their impact on detecting the segmentation failure. To evaluate the effect of dataset size, we trained the multi-branch network consisting of all features using the partial-dataset. Besides, we trained a single-branch network to examine the performance difference between single and multi-branch failure detection networks in \fsnet. Gray rows show the \fsnet\ accuracy when the failure detection network is multi-branched and trained using the entire dataset.}
\label{tbl:ablation_table}
\resizebox{\textwidth}{!}{%

\begin{tabular}{llll|llll|llll}

\\ \hline
\multirow{2}{*}{Dataset} &
  \multirow{2}{*}{\begin{tabular}[c]{@{}l@{}}Multi \\ Branch\end{tabular}} &
  \multirow{2}{*}{\begin{tabular}[c]{@{}l@{}}Full \\ Dataset\end{tabular}} &
  \multirow{2}{*}{\begin{tabular}[c]{@{}l@{}}Removed \\ Feature\end{tabular}} &
  \multicolumn{4}{c|}{FCN8} &
  \multicolumn{4}{c}{Deeplabv2} \\ \cline{5-12} 
                            &        &        &            & \small AP-Err$\uparrow$ & AP-Suc$\uparrow$ & AUC$\uparrow$   & FPR95$\downarrow$ & AP-Err$\uparrow$ & AP-Suc$\uparrow$ & AUC$\uparrow$   & FPR95$\downarrow$ \\ \hline
\multirow{9}{*}{Cityscapes} & \cmark & \cmark & $x$        & 60.01  & 98.78  & 92.83 & 24.37 & 52.17  & 99.08  & 93.17 & 22.85 \\
                            & \cmark & \cmark & $\mathcal{W}_{1}(l)$ & 51.40  & 98.04  & 89.37 & 39.42 & 45.53  & 98.34  & 89.49 & 42.44 \\
                            & \cmark & \cmark & $\mathcal{W}_{2}(l)$ & 61.88  & 98.73  & 92.69 & 25.24 & 50.51  & 99.00  & 92.55 & 24.63 \\
                            & \cmark & \cmark & $\mathcal{W}_{3}(l)$ & 67.20  & 98.96  & 94.17 & 21.99 & 53.19  & 99.08  & 93.25 & 23.24 \\
                            & \cmark & \cmark & $\mathcal{W}_{4}(l)$ & 45.85  & 96.11  & 79.85 & 62.31 & 36.19  & 96.50  & 76.67 & 64.90 \\
    \scriptsize in-distribution                        & \cmark & \cmark & $S_{E}(x)$ & 52.77  & 98.54  & 91.26 & 26.75 & 54.78  & 99.13  & 93.65 & 22.39 \\
    
                      \rowcolor{lightgray}    \cellcolor{white}  & 
                            
                             \cmark & \cmark & -          &
                            
                            67.83  & 98.98  & 94.35 & 21.52 & 57.84  & 99.13  & 93.97 & 21.39 \\
                            & \xmark & \cmark & -          & 65.92  & 98.99  & 94.37 & 20.29 & 54.96  & 99.14  & 93.63 & 22.63 \\
                            & \cmark & \xmark & -          & 65.48  & 98.90  & 93.83 & 23.59 & 50.64  & 99.03  & 92.70 & 26.05 \\ \hline
\multirow{9}{*}{BDD100k}    & \cmark & \cmark & $x$        & 67.75  & 85.36  & 80.59 & 55.15 & 70.09  & 91.74  & 86.71 & 43.83 \\
                            & \cmark & \cmark & $\mathcal{W}_{1}(l)$ & 62.38  & 82.01  & 75.55 & 64.44 & 67.10  & 91.10  & 85.18 & 51.26 \\
                            & \cmark & \cmark & $\mathcal{W}_{2}(l)$ & 67.02  & 85.39  & 80.74 & 54.43 & 67.86  & 91.38  & 85.73 & 45.48 \\
                            & \cmark & \cmark & $\mathcal{W}_{3}(l)$ & 71.81  & 86.38  & 82.83 & 52.73 & 69.96  & 91.15  & 86.18 & 47.32 \\
                            & \cmark & \cmark & $\mathcal{W}_{4}(l)$ & 65.37  & 85.24  & 79.09 & 58.70 & 65.03  & 90.58  & 82.80 & 56.50 \\
    \scriptsize out-distribution                        & \cmark & \cmark & $S_{E}(x)$ & 64.27  & 83.80  & 78.25 & 59.23 & 70.76  & 91.85  & 87.01 & 43.68 \\
                    \rowcolor{lightgray}    \cellcolor{white}        & 
                            \cmark & \cmark & -          & 72.29  & 86.63  & 83.17 & 51.81 & 71.81  & 91.76  & 86.94 & 43.56 \\
                            & \xmark & \cmark & -          & 69.63  & 85.27  & 81.12 & 55.51 & 68.14  & 90.74  & 85.32 & 48.29 \\
                            & \cmark & \xmark & -          & 72.00  & 85.67  & 83.05 & 52.95 & 66.67  & 90.67  & 84.87 & 49.92 \\ \hline
\multirow{9}{*}{Mapillary}  & \cmark & \cmark & $x$        & 65.08  & 93.75  & 87.12 & 39.00 & 59.45  & 95.40   & 87.66 & 40.54 \\
                            & \cmark & \cmark & $\mathcal{W}_{1}(l)$ & 59.57  & 91.25  & 82.68 & 51.93 & 56.46  & 95.87  & 87.41 & 41.71 \\
                            & \cmark & \cmark & $\mathcal{W}_{2}(l)$ & 59.86  & 92.68  & 84.92 & 43.41 & 56.26  & 95.16  & 86.60  & 43.24 \\
                            & \cmark & \cmark & $\mathcal{W}_{3}(l)$ & 67.68  & 93.78  & 87.57 & 40.60 & 58.61  & 94.97  & 86.84 & 44.04 \\
                            & \cmark & \cmark & $\mathcal{W}_{4}(l)$ & 61.18  & 93.30  & 84.13 & 48.67 & 53.69  & 95.78  & 84.90  & 47.48 \\
      \scriptsize out-distribution                      & \cmark & \cmark & $S_{E}(x)$ & 61.57  & 92.82  & 85.35 & 42.41 & 59.84  & 95.43  & 87.77 & 40.77 \\
                   \rowcolor{lightgray}    \cellcolor{white}         & 
                            \cmark & \cmark & -          & 68.29  & 93.76  & 87.75 & 40.19 & 63.19  & 95.68  & 88.95 & 37.36 \\
                            & \xmark & \cmark & -          & 66.22  & 92.99  & 86.83 & 41.37 & 58.70   & 95.38  & 87.45 & 41.52 \\
                            & \cmark & \xmark & -          & 68.03  & 92.62  & 86.53 & 44.08 & 54.35  & 94.37  & 84.89 & 49.02 \\ \hline
\end{tabular}%
}

\end{table*}

Based on experimental results, \fsnet\ outperforms SynthCP and other existing approaches. In the ablation study, we experimented with multiple configurations to find critical components of \fsnet. These configurations include single and multi-branch architecture, full and partial-dataset. In a single-branch setting, we used only a single encoder to extract features from the input and logits of the segmentation network. In a multi-branch (see Figure~\ref{fig:ral_architecture}), \fsnet\ used two encoders to extract features from the input and logits output. We tested how the dataset size impacts \fsnet\ using full-dataset and partial-dataset comprising $100\%$ and randomly selected $20\%$ Cityscapes training images to train \fsnet.

As described in the literature, SynthCP, Direct-prediction, and TCP train a segmentation network and apply that network on an unseen dataset to create a failure-dataset for failure detection training. Hence, the failure-dataset size is significantly smaller, which is $20\%$ - $25\%$ of the segmentation dataset. Therefore, these approaches can not take advantage of the entire available segmentation dataset. On the contrary, \fsnet\ introduces a joint architecture and simultaneously uses the full semantic segmentation dataset to train the segmentation and failure detection networks.  Table~\ref{tbl:ablation_table} shows the \fsnet\ accuracy for full-dataset and partial-dataset settings. In all cases, the full-dataset setting performs better than the partial-dataset setting.

Table~\ref{tbl:ablation_table} shows that the multi-branch network performs better than the single-branch network. This accuracy gain is possible by using separate encoders to extract more informative features from the image and logits output of the segmentation network.

In all existing approaches, either semantic segmentation or a separate network is used for failure detection. However, \fsnet\ failure detection network exploits multiple features from the segmentation network for failure detection. Figure~\ref{fig:ral_architecture} shows how these features are extracted from the segmentation network. We have removed one feature at a time from the \fsnet\ multi-branch failure detection network to study the impact of different features. In Table~\ref{tbl:ablation_table}, we have listed the feature name which is removed and the accuracy of \fsnet\ in all metrics after removing them. It shows that the \fsnet\ accuracy drops when any feature is removed from the failure detection network. Based on the  ablation study, the most significant features are $\mathcal{W}_{1}(l)$, $\mathcal{W}_{4}(l)$ and $S_{E}(x)$. Without these features, \fsnet\ failure detection accuracy drops below the SynthCP. Table~\ref{tbl:ablation_table} shows that our proposed joint architecture significantly improves the accuracy of \fsnet\ for detecting the failure of semantic segmentation network.

\section{Conclusion}
\label{sec:conclusion}
As deep learning-based semantic segmentation becomes an essential component for autonomous vehicles, identifying the segmentation failure is paramount for ensuring safety and robustness. This paper proposes a novel joint learning framework to train a semantic segmentation and corresponding failure detection network simultaneously. This failure detection network can identify the image area at pixel-level where the segmentation network has made an incorrect prediction. Therefore, our novel framework can inform downstream components in autonomous vehicle systems about expected semantic segmentation reliability. We show the effectiveness of \fsnet\ using multiple datasets, segmentation networks, and evaluation metrics.
\bibliographystyle{IEEEtran}
\bibliography{ref}
\end{document}